\documentclass[10pt,final,journal]{IEEEtran}
\usepackage{amssymb}
\usepackage{amsfonts}
\usepackage[english]{babel}
\usepackage[autostyle]{csquotes}
\usepackage{amsbsy}

\usepackage{multirow}
\usepackage{multicol}
\usepackage{subfigure}
\usepackage{textcomp}
\usepackage[export]{adjustbox}
\usepackage{balance}
\usepackage{xcolor}
\usepackage{array}
\usepackage{longtable}
\usepackage{calc}
\usepackage{url}
\usepackage{amssymb}
\usepackage{gensymb}
\usepackage{afterpage}
\usepackage{forest}
\usepackage{caption}
\usepackage{stackengine}
\usepackage{lipsum}
\usepackage{mathtools}
\usepackage[utf8]{inputenc}
\usepackage[T1]{fontenc}
\usepackage{epstopdf}
\usepackage{tabularx,booktabs}
\usepackage{booktabs,ragged2e}
\usepackage[flushleft]{threeparttable}
\usepackage{cuted}
\usepackage{cite}
\usepackage{amsmath}
\usepackage{algorithmic}
\usepackage{array}
\usepackage{graphics}
 \usepackage[caption=false,font=normalsize,labelfont=sf,textfont=sf]{subfig}
 \usepackage{dblfloatfix}
\usepackage{url}
\makeatletter

\begin{document}
\title{Anomaly detection in non-stationary videos using time-recursive differencing network based prediction}
\author{Gargi~V.~Pillai
        and~Debashis~Sen$^*$,~\IEEEmembership{Senior Member,~IEEE}
\thanks{G. V. Pillai and D. Sen are with the Department of Electronics and Electrical Communication Engineering, Indian Institute of Technology, Kharagpur, India, 721302. $^*$E-mail: dsen@ece.iitkgp.ac.in}}

\markboth{IEEE GEOSCIENCE AND REMOTE SENSING LETTERS}
{Shell \MakeLowercase{\textit{et al.}}: Bare Demo of IEEEtran.cls for Journals}
\maketitle
\begin{abstract}
Most videos, including those captured through  aerial remote sensing, are usually non-stationary in nature having time-varying feature statistics. Although, sophisticated reconstruction and prediction models exist for video anomaly detection, effective handling of non-stationarity has seldom been considered explicitly.
In this paper, we propose to perform prediction using a time-recursive differencing network followed by autoregressive moving average estimation for video anomaly detection. The differencing network is employed to effectively handle non-stationarity in video data during the anomaly detection. Focusing on the prediction process, the effectiveness of the proposed approach is demonstrated considering a simple optical flow based video feature, and by generating qualitative and quantitative results on three aerial video datasets and two standard anomaly detection video datasets. EER, AUC and ROC curve based comparison with several existing methods including the state-of-the-art reveal the superiority of the proposed approach.
\end{abstract}

%
\IEEEpeerreviewmaketitle
\section{Introduction}
\label{sec:intro}
In today\textquotesingle s world, security has become an important part of one\textquotesingle s  life. Unmanned aerial vehicles (UAVs) are being increasingly used for various surveillance applications \cite{mesquita2019fully}. Therefore, demand for intelligent video surveillance devices that detect unusual behavior automatically is growing. Anomaly detection is the identification of abnormal patterns in data that are by definition unusual or rare events, where events are change inducing actions or combinations of actions \cite{chandola2009anomaly}.
\par In video anomaly detection (VAD), there are two main parts; appropriate feature extraction for event representation, and model to evaluate if the observed activity is anomalous or not. Hand-crafted features extracted for detecting anomaly can be directly based on optical flow like motion information, histogram of optical flow, and motion boundary histograms \cite{zhu2018real,leyva2017video}, or can be based on the extraction of local patterns, histogram of gradients\cite{xiao2015learning} and spatio–temporal gradients\cite{xu2014video}. Feature extraction can also be learned from available data (anomalous or not), which has been performed by the recent VAD techniques based on deep learning networks such as auto-encoders\cite{sabokrou2015real,xu2015learning}, convolutional neural networks\cite{ionescu2019object}, long short term memory networks \cite{nawaratne2019spatiotemporal}. This paper focuses on the model to classify the observed event as an anomaly or not, irrespective of the underlying features used. The classification is mainly based on analyzing the discrepancy between the expected and the actual data.
\begin{figure*}[h]
\centering
\captionsetup{justification=centering}
\setlength{\fboxsep}{0.008pt}%
			\setlength{\fboxrule}{0.8pt}%
			\includegraphics[width=16cm, height=4 cm]{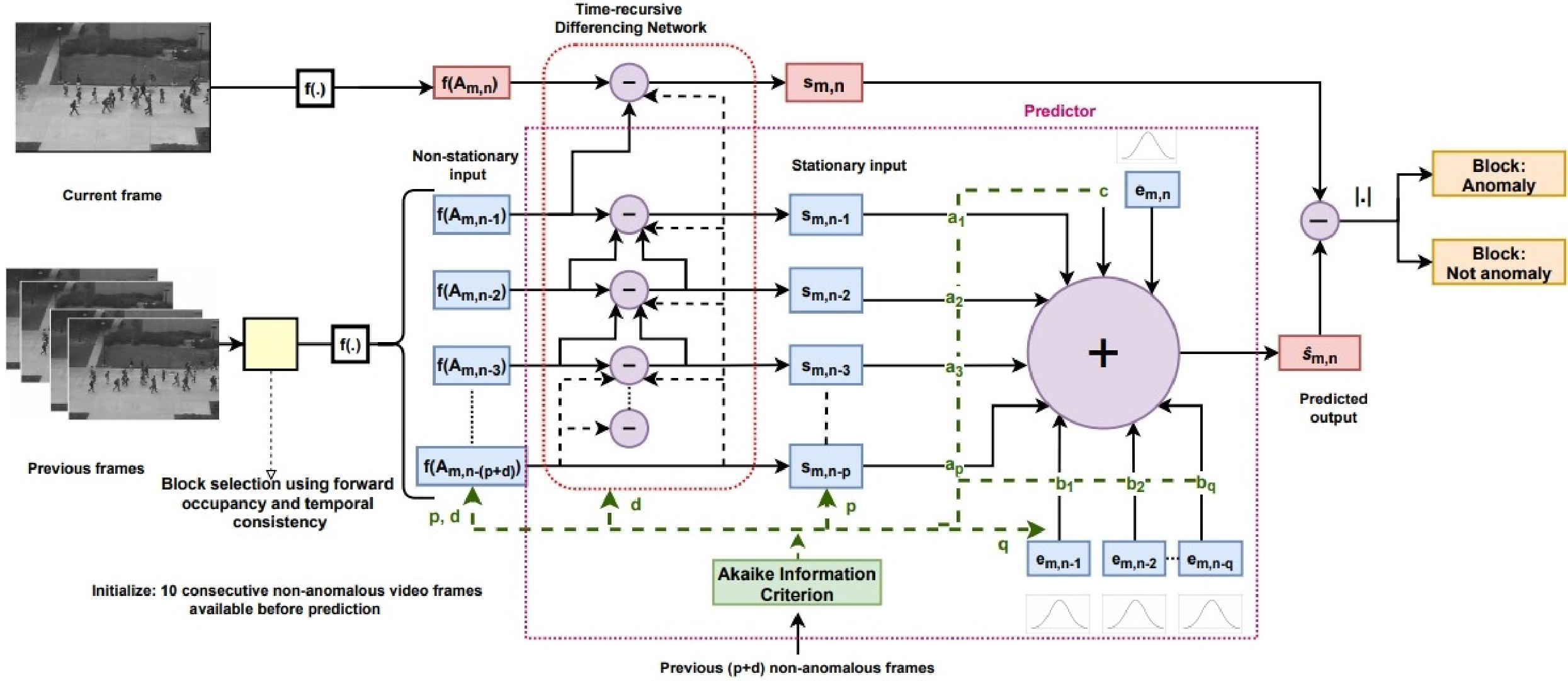}
\caption{Schematic of the proposed approach, where the time-recursive differencing module handles non-stationary inputs to produce stationary inputs for anomaly detection. $f(A_{m,n})$ refers to feature extracted from the $m^{th}$ block of $n^{th}$ frame.}
\label{bd}
\end{figure*}
\par In literature, different approaches have been proposed for modeling anomaly through the said discrepancy analysis. For VAD, reconstruction models consider the discrepancy between the actual frame and its reconstruction from the known/learnt model of non-anomalous frame. Most approaches that use autoencoder and its variants \cite{zhu2018real, sabokrou2015real} trained on non-anomalous data are such examples. Prediction models predict the current frame from multiple previous frames and consider the discrepancy between the predicted and the actual frames. Auto-regressive models \cite{abati2019latent}, convolutional Long-Short-Term-Memory models \cite{nawaratne2019spatiotemporal} and Generative Adversarial Networks \cite{ravanbakhsh2017abnormal} are such examples. Recently, models combining prediction and reconstruction strategies have been attempted \cite{abati2019latent}.
\par Although advanced classification models and sophisticated features have been used for VAD, the non-stationary nature of video data (extracted video features) has not been explicitly considered (see Section~\ref{sec:mot}). An aspect of non-stationarity in videos that will evidently affect VAD is the time-varying nature of feature statistics. Videos captured using aerial remote sensing \cite{fu2020object} are usually non-stationary with such time-varying statistics. Hence, mathematical incorporation of non-stationarity that allows the classifier to model time-changing patterns is expected to improve anomaly detection performance in all videos, including those obtained aerially.
\par In this paper, we propose the use of a prediction model for anomaly detection in videos that handles non-stationary video data through differencing. The concept of using a time-recursive differencing network followed by estimation based on autoregression and moving average of regression errors is considered in our approach. Our model is generic in nature, which can be used in conjunction with any sophisticated feature. Video frames are spatially divided into non-overlapping blocks and our prediction model is applied on these temporally varying blocks. The comparison of the predicted feature of a block with the actual to detect anomaly is locally adapted, where spatio-temporal consistency is also maintained. We evaluate our approach both on aerial video datasets and standard anomaly detection datasets. Experimental results obtained using a simple optical flow based feature qualitatively and quantitatively demonstrates the effectiveness of our proposal of using time-recursive differencing network for video anomaly detection. Further, comparison of the quantitative performance with relevant existing approaches demonstrates the superiority of our approach. We choose a simple feature in our VAD approach in order to focus on the performance of our prediction method, and use of relevant sophisticated features shall be explored in future. The main contributions of this paper are:
\begin{itemize}
    \item It introduces the use of time differencing, a concept borrowed from time series analysis, for anomaly detection in non-stationary videos and finds it immensely effective.
    \item It presents a time-recursive differencing network form, which allows higher order recurrence driven processing of video frames.
\end{itemize}
\par Our approach is introduced in Section~\ref{sec:approach}, and its result analysis is given in Section~\ref{sec:result}. Section~\ref{sec:conclusion} concludes the paper.
\section{Proposed Prediction Approach for Video Anomaly Detection}
\label{sec:approach}
Pictorial overview of our proposed VAD approach is shown in Fig.~\ref{bd}. As mentioned in Section~\ref{sec:intro}, video frames are spatially divided into non-overlapping blocks, and our prediction is performed on these temporally varying blocks. Any arbitrary feature can be extracted from these blocks for use in our prediction. The feature we consider in our experiments is mentioned in Section~\ref{sec:result}.
\subsection{Motivation}
\label{sec:mot}
Our primary goal is to perform the prediction in such a way that the non-stationarity in the temporally varying blocks is handled appropriately. As mentioned in \cite{adhikari2013introductory,IEEEexample:brockwell2016introduction}, non-stationarity in time series data can be effectively handled using time-recursive differencing as performed in auto regressive integrated moving average modeling. Most state-of-the-art approaches use neural networks, recurrent or not, at most of second-order (like LSTM\cite{nawaratne2019spatiotemporal}), to perform prediction for VAD. Such networks are very effective in learning invariance to local deformations in data \cite{wiatowski2017mathematical}. However, it is obvious that non-recurrent neural networks do not have the provision to learn time-recursive differencing, and first and second-order recurrent neural networks can not learn higher-order time-recursive differencing \cite{zhang2017does, monner2012generalized}. Depending on the video data at hand, effective handling of non-stationarity might require higher-order time-recursive differencing, which we achieve through our time-recursive difference network.
\subsection{Time-recursive Differencing based Anomaly Detection}
\label{sec:method}
We consider that $F$ (e.g. 10) consecutive video frames having no anomaly is available before our prediction starts for estimating certain parameters as mentioned in Section~\ref{sec:param}.
\par In our prediction model, each frame is divided into $N \times N$ blocks. Consider the $j^{th}$ block in the $i^{th}$ frame. KNN matting \cite{chen2013knn} based foreground-background segregation is used to binarize the block yielding $\beta_{j,i}$, which is used to compute forward occupancy $o_{j,i}$ as follows:
  \begin{flalign}
o_{j,i} = \small{\frac{1}{N^2} \sum_{\alpha_1=1}^{N}\sum_{\alpha_2=1}^{N}\beta_{j,i}(\alpha_1,\alpha_2),\ \beta_{j,i}(\alpha_1,\alpha_2) \in \{0,1\}}
\end{flalign}
\par The forward occupancy value is the total number of foreground pixels in the block. The blocks with $o_{j,i} \neq 0$, which represent some part of the foreground, are then subjected to temporal consistency analysis. Temporal consistency is achieved by considering only those blocks with some foreground part, which have such blocks in their immediate temporal neighborhood. So, features are extracted only from these active blocks, using which prediction is performed. One may additionally consider only those active blocks for prediction where the maximum of the feature values is above/below a certain value $\lambda_{f}$ to handle background noise. Since the above operations will eliminate some blocks of certain frames from use in prediction, we refer an active block which is the $m^{th}$ remaining block in the $n^{th}$ remaining frame as $A_{m,n}$.
\par  To decide if $A_{m,n}$ is anomalous or not, we perform prediction using all the $m^{th}$ blocks in previous $p+d$ frames, that is, $A_{m,k'},\ k'=n-1, n-2,\cdots, n-(p+d)$, where $p$ is determined as explained in Section~\ref{sec:param} and $d$ is explained later. Note that, video frames are subjected to anomaly detection sequentially, and the blocks considered from the previous $p+d$ frames are taken such that they are not represented by anomalous features. We refer the feature representing or extracted from $A_{m,k'}$ as $f(A_{m,k'})$. The first operation performed for the prediction is to generate a stationary signal $s_{m,k}$ from $f(A_{m,k})$, where $k=n-1, n-2,\cdots, n-p$. This is achieved through $d^{th}$-order time-recursive differencing performed by the relevant network shown in Fig.~\ref{bd}. Consider the $2^{nd}$-order differencing below:
\begin{flalign}
    (f(A_{m,k})-&f(A_{m,k-1}))-(f(A_{m,k-1})-f(A_{m,k-2}))\nonumber \\
    &=f(A_{m,k})-2f(A_{m,k-1})+f(A_{m,k-2})\nonumber \\
    &=(1-\Lambda)^2f(A_{m,k})\qquad
\end{flalign}
where $\Lambda^z(f(A_{m,k}))=f(A_{m,k-z})$. Therefore, the $d^{th}$-order time-recursive differencing network implements:
\begin{flalign}
   s_{m,k}=(1-\Lambda)^{d}f(A_{m,k})
\end{flalign}
where $d$ is determined as explained in Section~\ref{sec:param}. Such time-recursive differencing increases the overlap among the inputs generating the samples $s_{m,k}$. The said differencing has been shown to reduce discrepancy in the local first-order moment of a non-stationary signal through the overlap increase along the dimension where the differencing is performed \cite{pankratz2009forecasting}. Higher the order of differencing, more is the discrepancy reduction, and hence, the time-recursive differencing leads to a first-order stationary output $s_{m,k}$ \cite{IEEEexample:brockwell2016introduction}.
\begin{figure}[h]
		\centering
		\begin{tabular}{ c c}
			\setlength{\fboxsep}{0.008pt}%
			\setlength{\fboxrule}{0.8pt}%
			\includegraphics[width=3.5 cm, height=.5 cm]{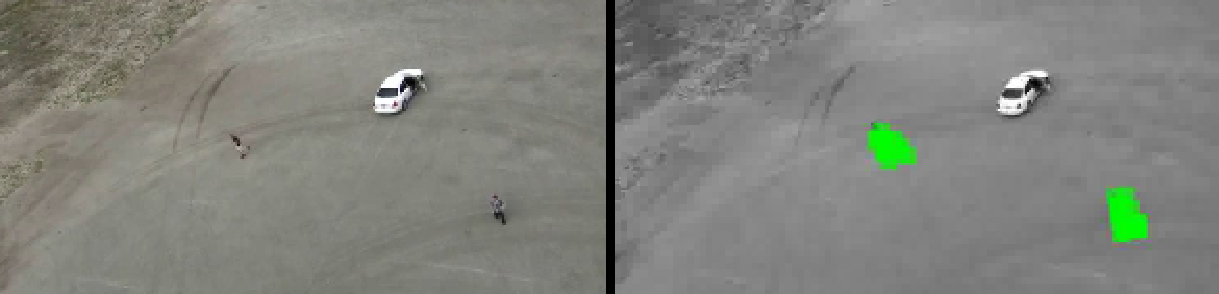} &
				\setlength{\fboxsep}{0.008pt}%
			\setlength{\fboxrule}{0.8pt}%
		\includegraphics[width=3.5 cm, height=.5 cm]{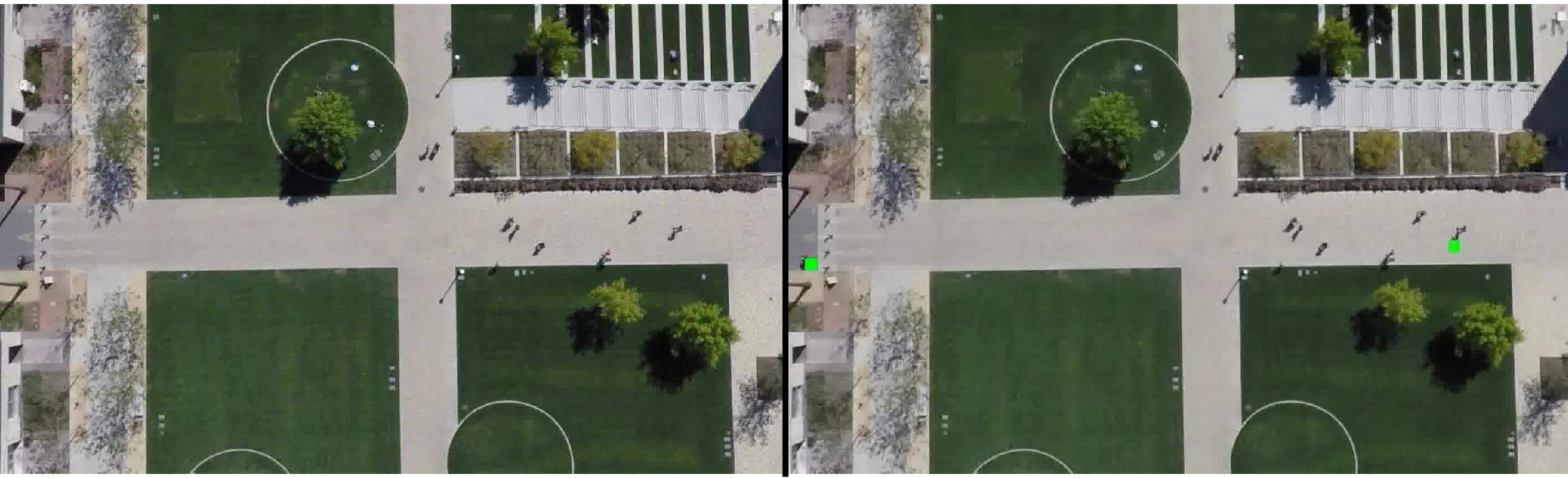}
			\\
		\small{(a)}&\small{(b)}
		\end{tabular}
		\caption{Anomalies detected in frames from aerial videos, (a) walking people suddenly start running in a video from the UCF-AA dataset and (b) a cyclist appears in the scene in a video from the SD dataset}
		\label{result_frames_aerial}
	\end{figure}
	\begin{figure}[h]
		\centering
		\begin{tabular}{ c c}
			\setlength{\fboxsep}{0.008pt}%
			\setlength{\fboxrule}{0.8pt}%
			\includegraphics[width=3.5 cm, height=.5 cm]{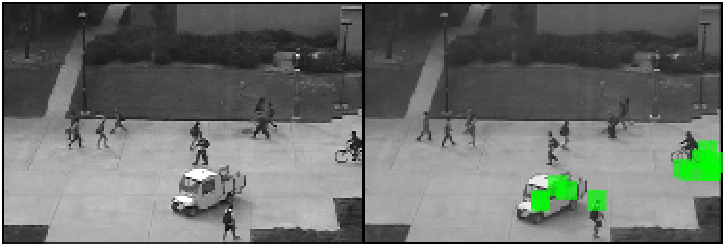} &
				\setlength{\fboxsep}{0.008pt}%
			\setlength{\fboxrule}{0.8pt}%
		\includegraphics[width=3.5 cm, height=.5 cm]{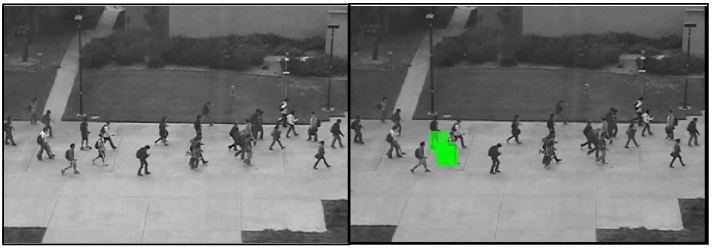} 
			\\
	\small{(a)}&\small{(b)}
		\end{tabular}
		\caption{Anomalies detected in frames from the UCSD Ped2 dataset, (a) vehicle and (b) skateboarding in pedestrian path}
		\label{result_frames}
	\end{figure}
	\begin{figure}[h]
		\centering
		\begin{tabular}{p{.1cm} c c c}
\raisebox{0.2 em}{\small{(a)}}&	
			\setlength{\fboxsep}{0.008pt}%
			\setlength{\fboxrule}{0.8pt}%
			\fbox{\includegraphics[width=1.5 cm, height=.4 cm]{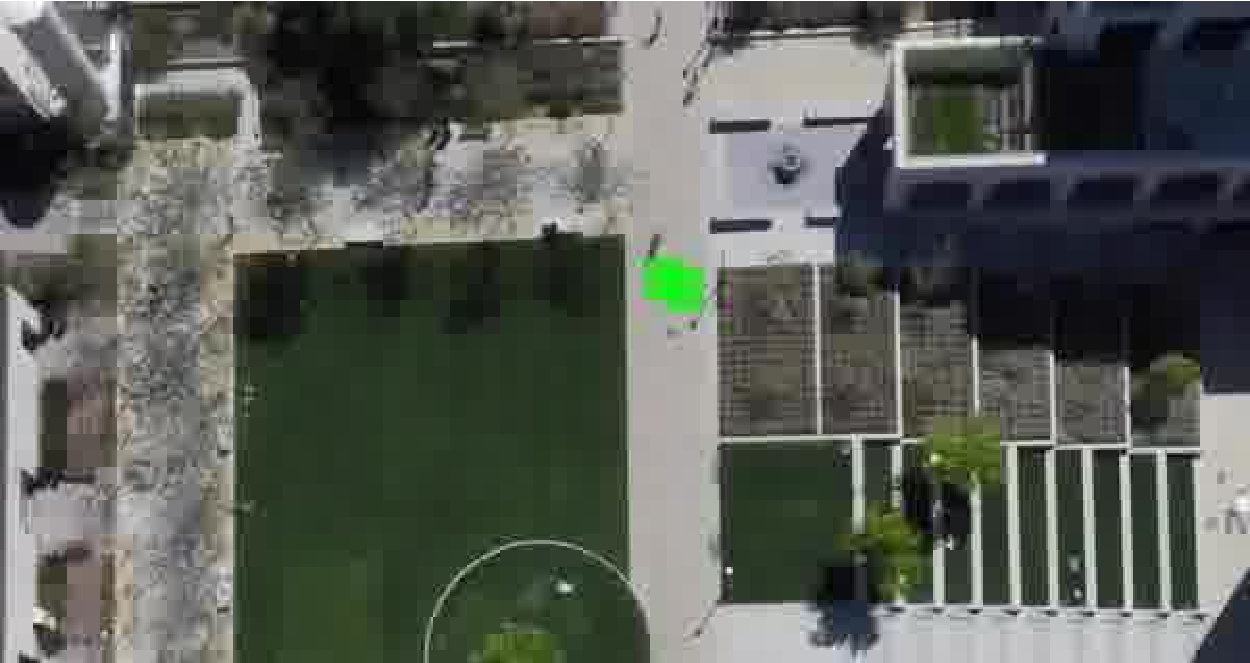}} &
				\setlength{\fboxsep}{0.008pt}%
			\setlength{\fboxrule}{0.8pt}%
			\fbox{\includegraphics[width=1.5 cm, height=.4 cm]{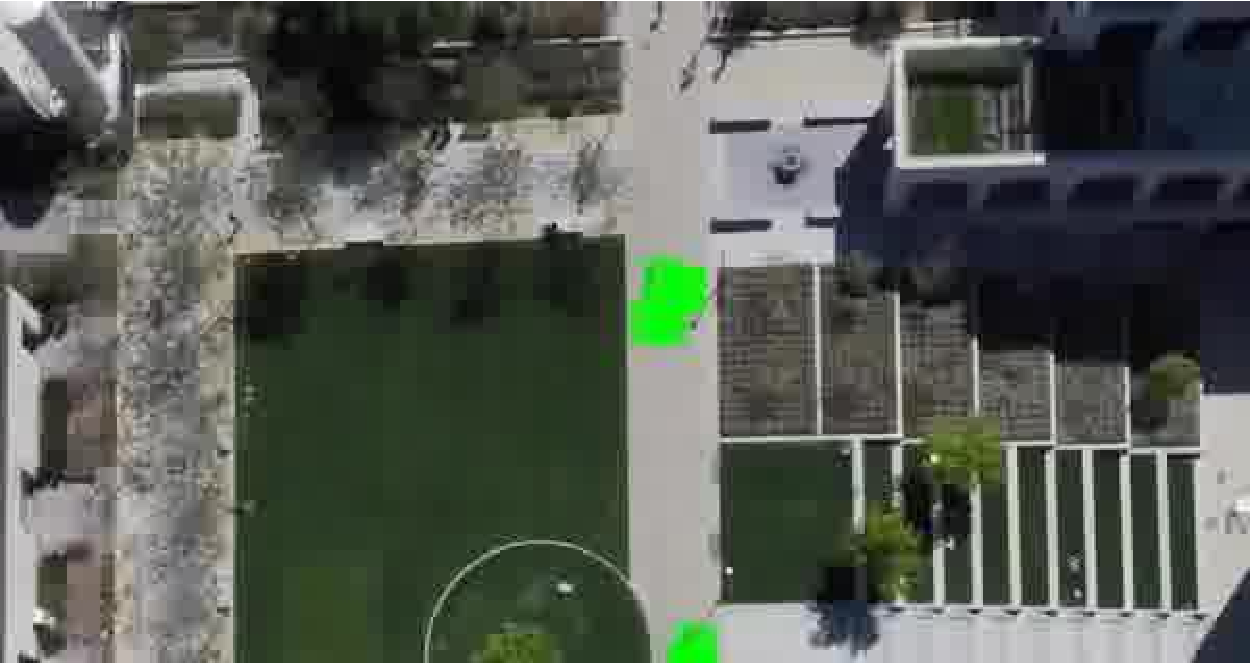}}&
				\setlength{\fboxsep}{0.008pt}%
			\setlength{\fboxrule}{0.8pt}%
			\fbox{\includegraphics[width=1.5 cm, height=.4 cm]{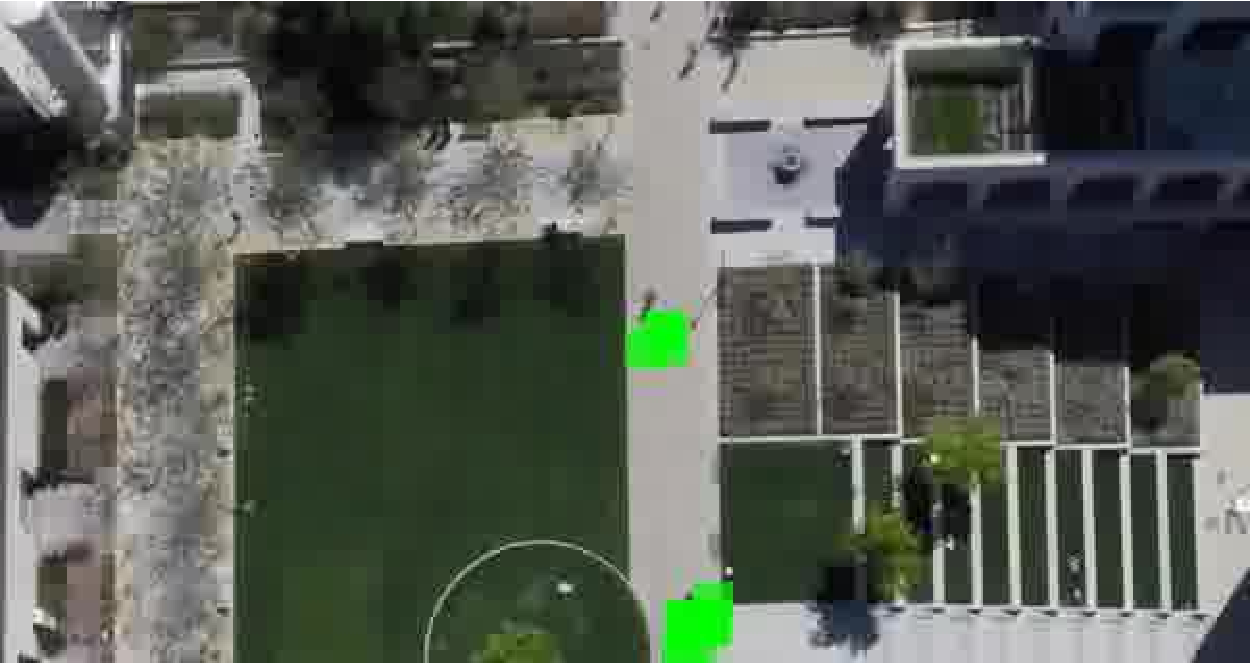}}
		\\
		\raisebox{0.2 em}{\small{(b)}}&	
			\setlength{\fboxsep}{0.008pt}%
			\setlength{\fboxrule}{0.8pt}%
			\fbox{\includegraphics[width=1.5 cm, height=.4 cm]{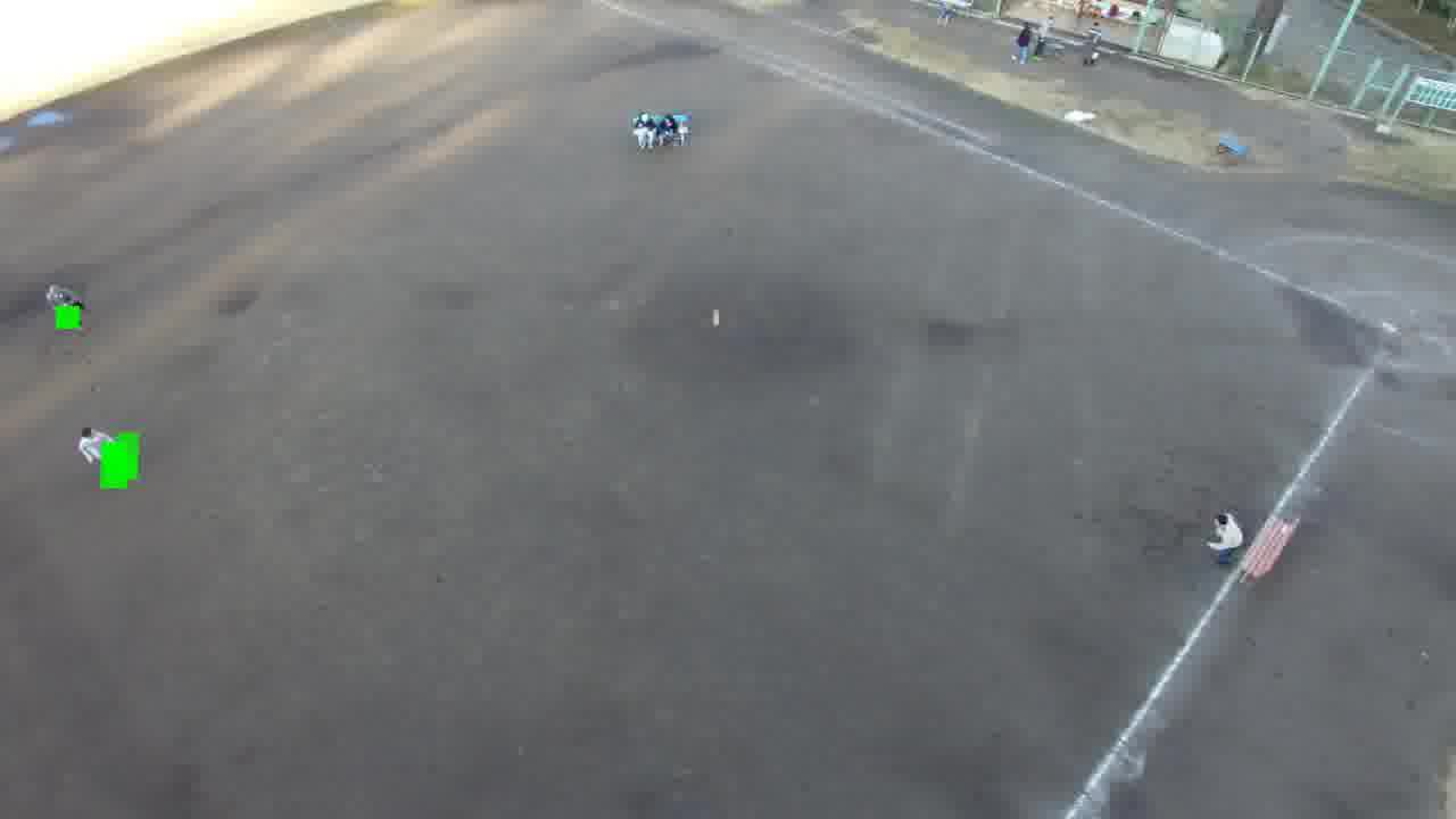}} &
				\setlength{\fboxsep}{0.008pt}%
			\setlength{\fboxrule}{0.8pt}%
			\fbox{\includegraphics[width=1.5 cm, height=.4cm]{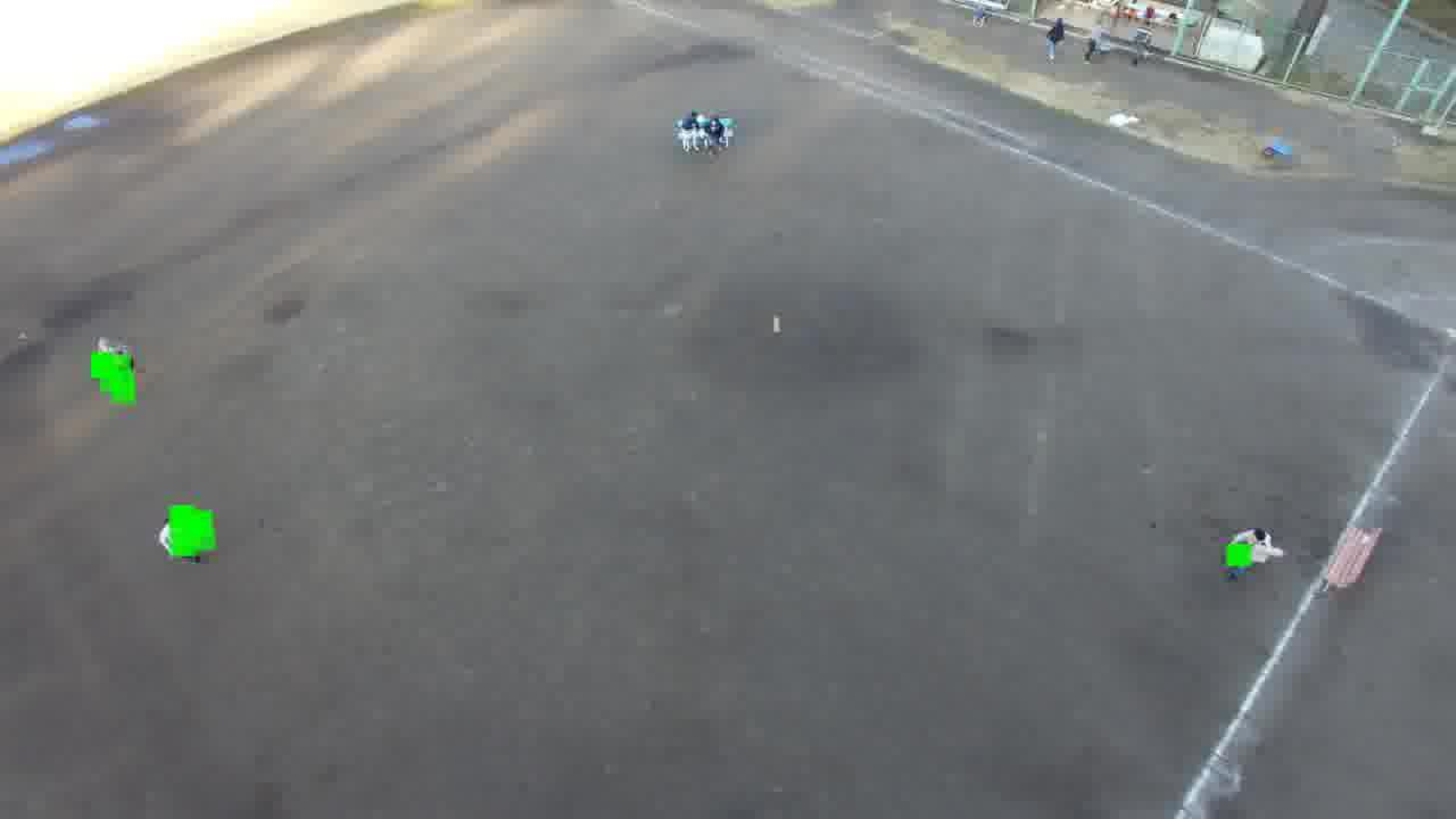}}&
				\setlength{\fboxsep}{0.008pt}%
			\setlength{\fboxrule}{0.8pt}%
			\fbox{\includegraphics[width=1.5 cm, height=.4 cm]{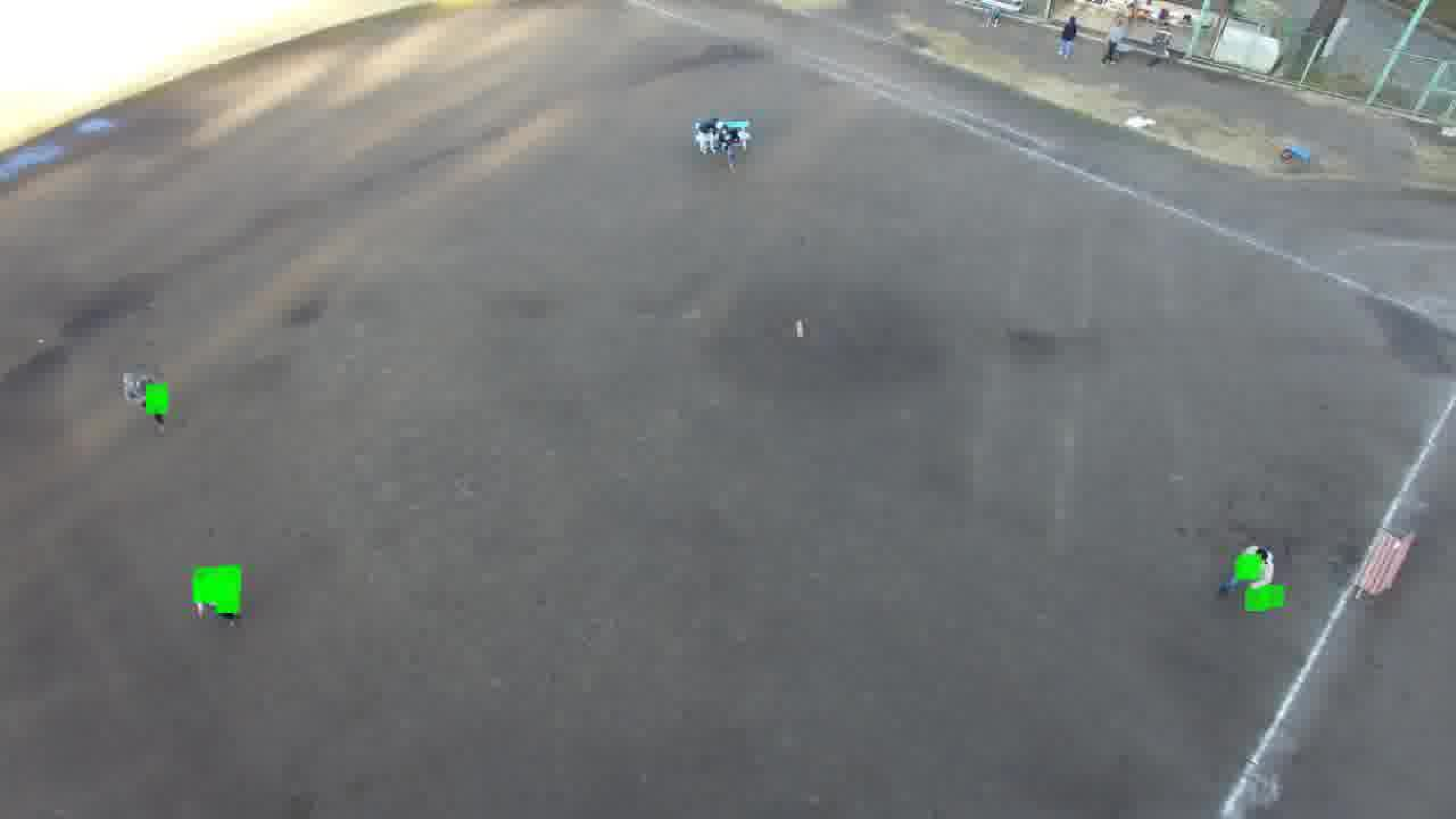}}
			\\
		\raisebox{0.2 em}{\small{(c)}}&	
			\setlength{\fboxsep}{0.008pt}%
			\setlength{\fboxrule}{0.8pt}%
			\fbox{\includegraphics[width=1.5 cm, height=.4 cm]{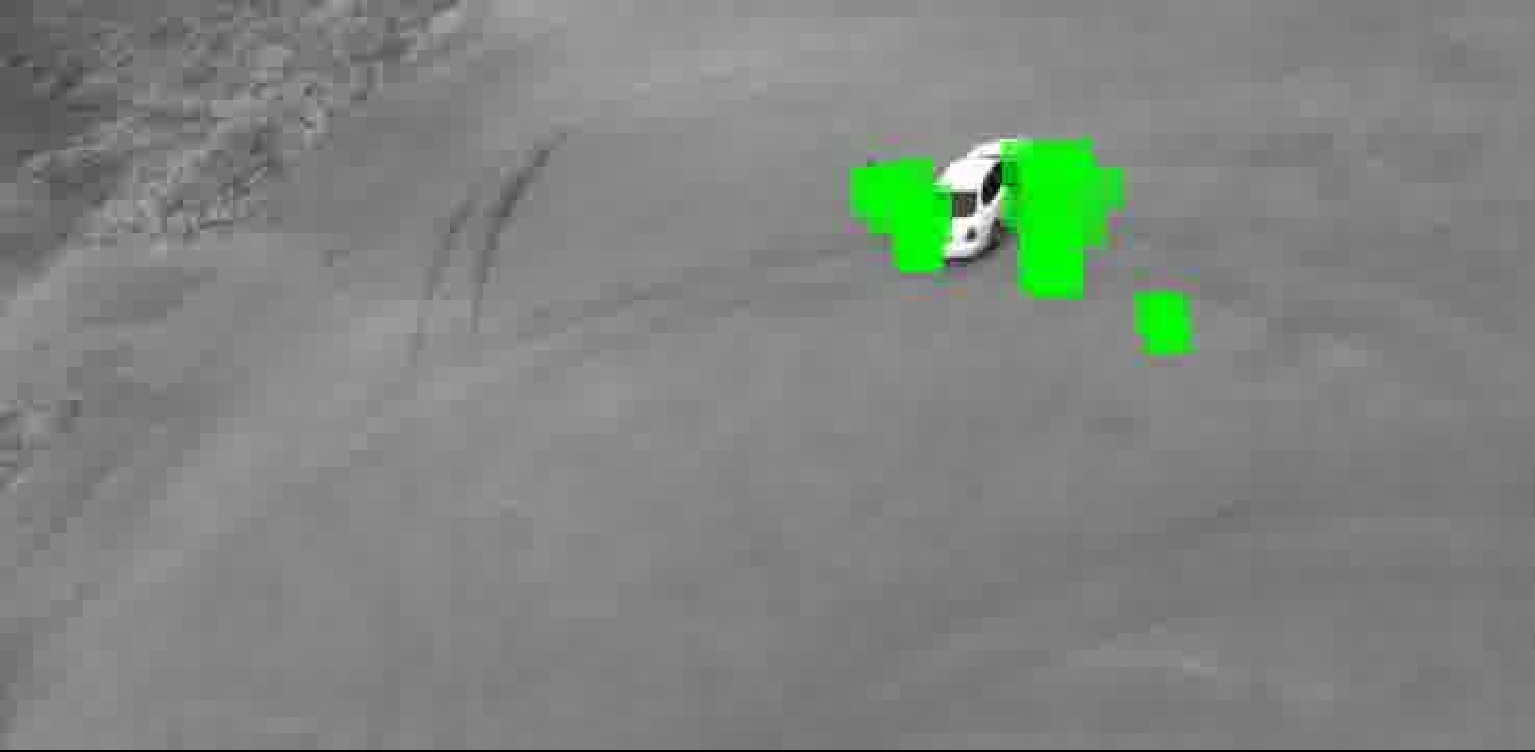}} &
				\setlength{\fboxsep}{0.008pt}%
			\setlength{\fboxrule}{0.8pt}%
			\fbox{\includegraphics[width=1.5 cm, height=.4 cm]{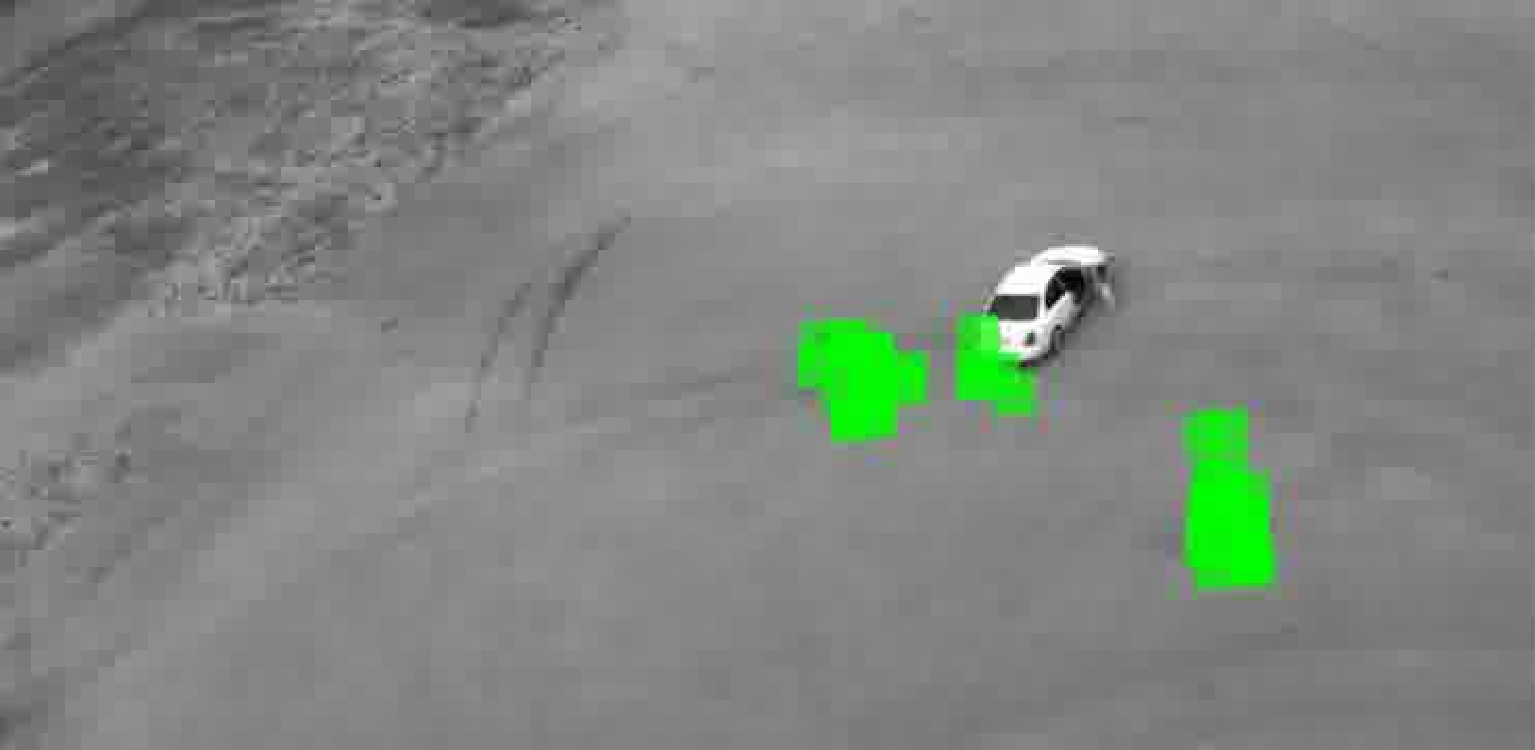}}&
				\setlength{\fboxsep}{0.008pt}%
			\setlength{\fboxrule}{0.8pt}%
			\fbox{\includegraphics[width=1.5 cm, height=.4 cm]{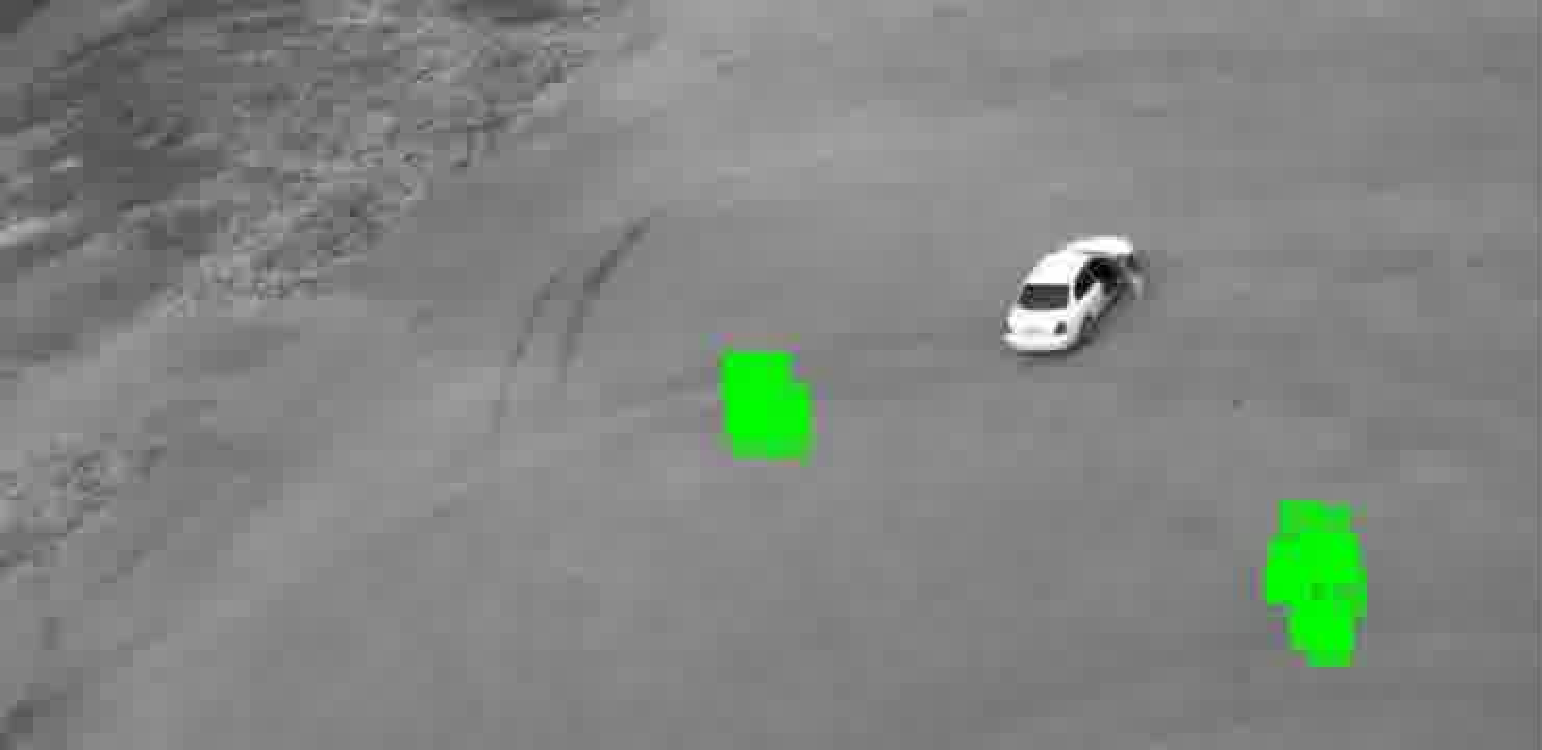}}
		\\
\raisebox{0.2 em}{\small{(d)}}&
	\setlength{\fboxsep}{0.008pt}%
			\setlength{\fboxrule}{0.8pt}%
			\fbox{\includegraphics[width=1.5 cm, height=.4 cm]{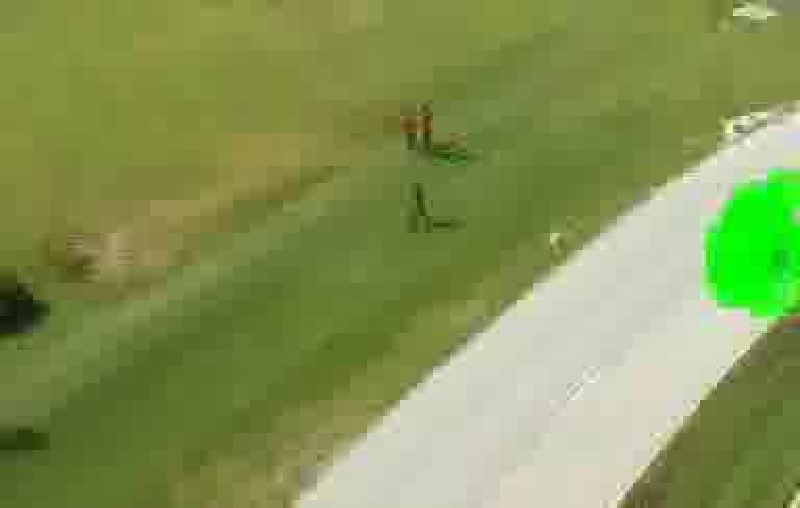}} &
				\setlength{\fboxsep}{0.008pt}%
			\setlength{\fboxrule}{0.8pt}%
			\fbox{\includegraphics[width=1.5 cm, height=.4 cm]{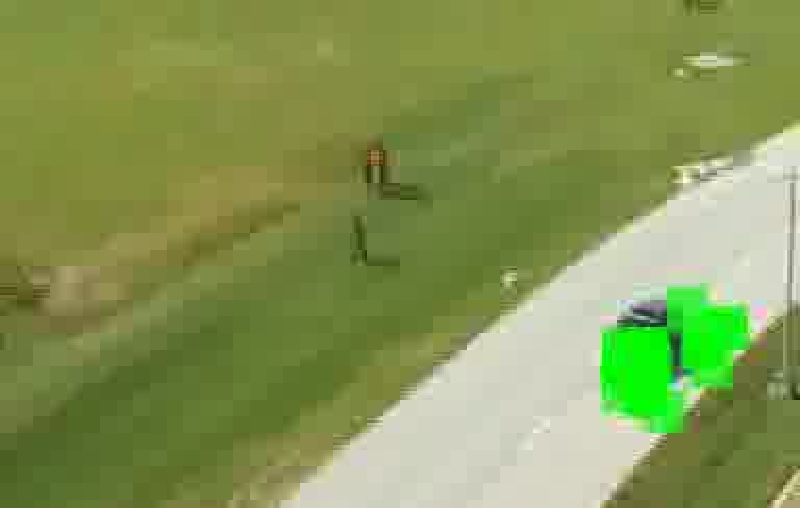}}&
				\setlength{\fboxsep}{0.008pt}%
			\setlength{\fboxrule}{0.8pt}%
			\fbox{\includegraphics[width=1.5 cm, height=.4 cm]{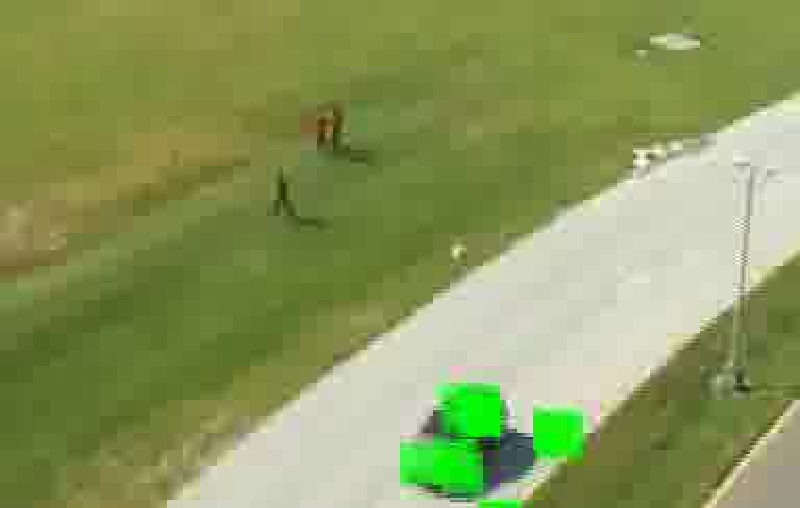}}
			\\
			\raisebox{0.2 em}{\small{(e)}}&
\setlength{\fboxsep}{0.008pt}%
			\setlength{\fboxrule}{0.8pt}%
			\fbox{\includegraphics[width=1.5 cm, height=.4 cm]{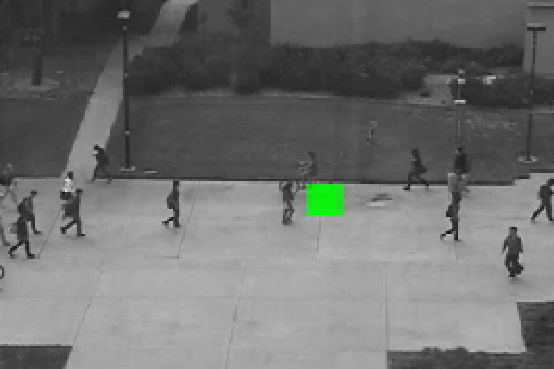}} &
				\setlength{\fboxsep}{0.008pt}%
			\setlength{\fboxrule}{0.8pt}%
			\fbox{\includegraphics[width=1.5 cm, height=.4 cm]{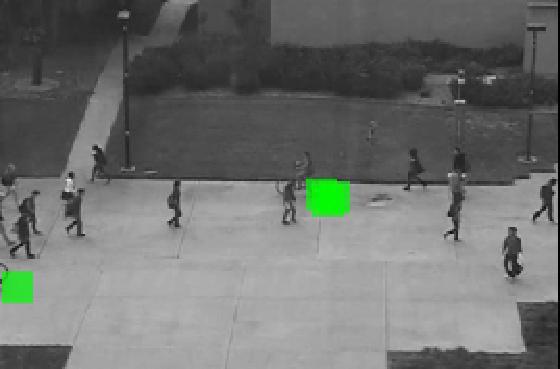}}&
				\setlength{\fboxsep}{0.008pt}%
			\setlength{\fboxrule}{0.8pt}%
			\fbox{\includegraphics[width=1.5 cm, height=.4 cm]{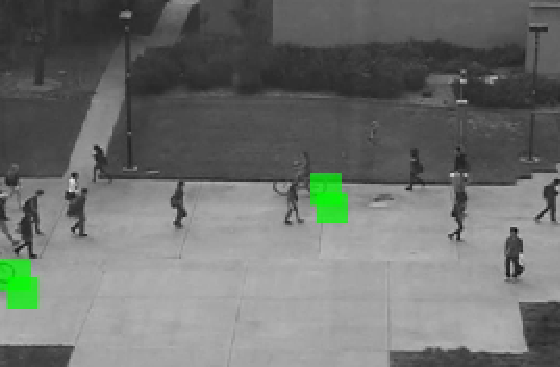}}
			\\
			\raisebox{0.2 em}{\small{(f)}}&
	\setlength{\fboxsep}{0.008pt}%
			\setlength{\fboxrule}{0.8pt}%
		\fbox{\includegraphics[width=1.5 cm, height=.4 cm]{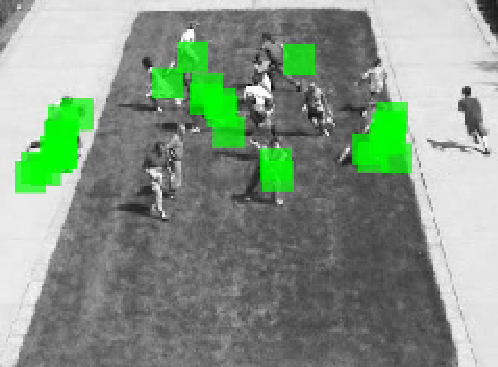}} &
				\setlength{\fboxsep}{0.008pt}%
			\setlength{\fboxrule}{0.8pt}%
			\fbox{\includegraphics[width=1.5 cm, height=.4 cm]{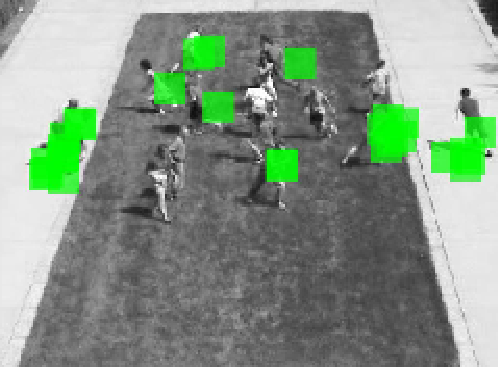}}&
				\setlength{\fboxsep}{0.008pt}%
			\setlength{\fboxrule}{0.8pt}%
			\fbox{\includegraphics[width=1.5 cm, height=.4 cm]{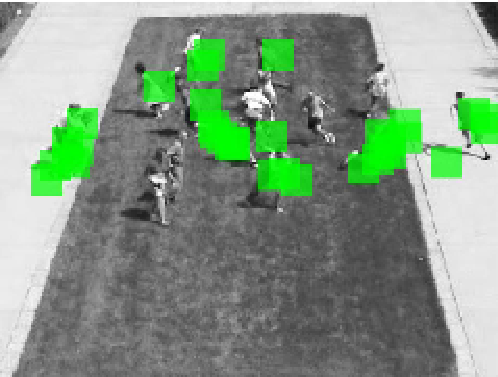}}
			\\
		\raisebox{0.2 em}{\small{(g)}}&
	\setlength{\fboxsep}{0.008pt}%
			\setlength{\fboxrule}{0.8pt}%
			\fbox{\includegraphics[width=1.5 cm, height=.4 cm]{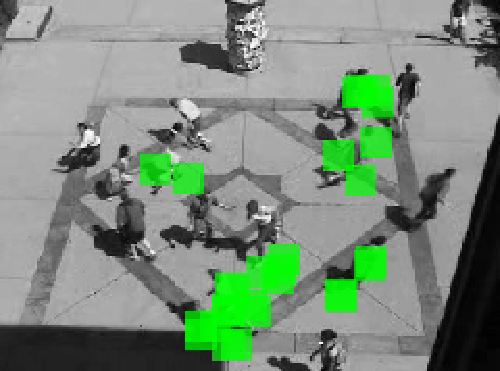}} &
				\setlength{\fboxsep}{0.008pt}%
			\setlength{\fboxrule}{0.8pt}%
			\fbox{\includegraphics[width=1.5 cm, height=.4 cm]{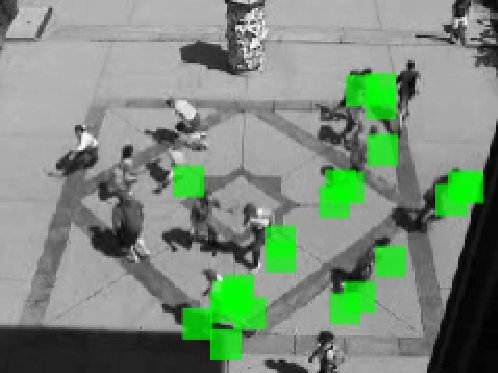}}&
				\setlength{\fboxsep}{0.008pt}%
			\setlength{\fboxrule}{0.8pt}%
			\fbox{\includegraphics[width=1.5 cm, height=.4 cm]{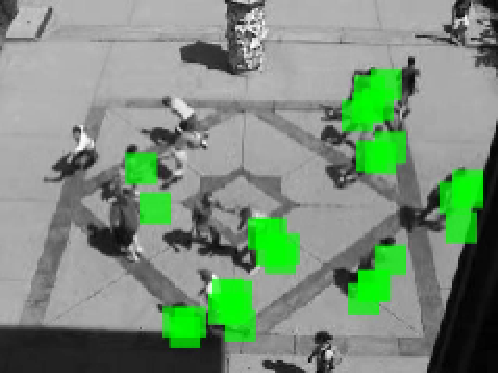}} 
			\\
          \raisebox{0.2 em}{\small{(h)}}&
        	\setlength{\fboxsep}{0.008pt}%
			\setlength{\fboxrule}{0.8pt}%
			\fbox{\includegraphics[width=1.5 cm, height=.4 cm]{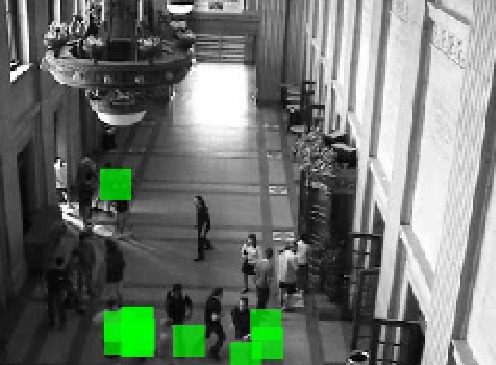}} &
				\setlength{\fboxsep}{0.008pt}%
			\setlength{\fboxrule}{0.8pt}%
			\fbox{\includegraphics[width=1.5 cm, height=.4 cm]{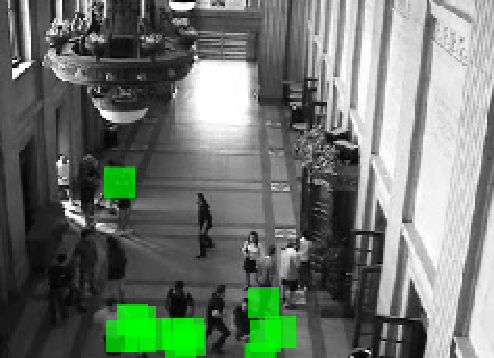}}&
				\setlength{\fboxsep}{0.008pt}%
			\setlength{\fboxrule}{0.8pt}%
			\fbox{\includegraphics[width=1.5 cm, height=.4 cm]{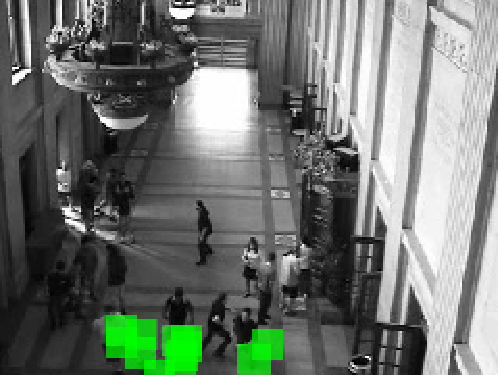}} 
			\\
		\end{tabular}
		\caption{Anomalies detected in consecutive frames of videos from (a) SD, (b) OA, (c)-(d) UCF-AA, (e) UCSD Ped2 and (f)-(h) UMN datasets}
		\label{consecutive frames}
	\end{figure}
\par $s_{m,k}, \forall k$ is then used to get $\hat{s}_{m,n}$, the predicted value of $s_{m,n}$, which is the stationary signal obtained from $f(A_{m,n})$ through the time-recursive differencing network. Autoregression of $s_{m,k}$ and moving average of corresponding regression errors as white noise is considered to get $\hat{s}_{m,n}$ as follows:
\begin{eqnarray}\nonumber
 \hat{s}_{m,n} = a_{1}s_{m,n-1} + \cdots + a_{p} s_{m,n-p} + b_{1}e_{m,n-1}\\ + \cdots +
\left. b_{q}e_{m,n-q} + e_{m,n}+ c \right.
\label{eqn}
\end{eqnarray}
where model coefficients $a_{1\text{---}p}, b_{1\text{---}q},$ and $c$ are obtained as stated in Section~\ref{sec:param}, and 
  \begin{align}
    e_{m,w}=\sigma Z_{m,w},\quad Z_{m,w}\sim \mathcal{N}(0,1)
    \end{align}
so that $e_{m,w} \sim \mathcal{N}(0,\sigma^{2})$. In the above, $Z_{m,w}$ is i.i.d. and $w=n, n-1, n-2,\cdots, n-q$, where $q$ is determined as explained in Section~\ref{sec:param}. 
Given the estimate $\hat{s}_{m,n}$, the $m^{th}$ block of $n^{th}$ frame is considered anomalous if $|s_{m,n} - \hat{s}_{m,n}|>\lambda_A$.  
Finally, local spatial consistency is imposed by considering only those blocks anomalous, which have such blocks in their immediate spatial neighborhood. Frame-level and pixel-level video anomaly can be directly derived from the above.
\subsection{Model and Hyper-Parameters}
\label{sec:param}
Let the $F$ consecutive non-anomalous video frames available before prediction be $V_x, x=1,\cdots,F$. Initial estimate of the hyper-parameters $p,q,$ and $d$, and the model coefficients $a_{1\text{---}p}, b_{1\text{---}q},$ and $c$ are obtained through the Akaike information criterion (AIC) \cite{IEEEexample:brockwell2016introduction}, which involves computation of the likelihood (probability) $P(S_{F}/\theta, S_{x}), \forall x\ \& \ x\ne F$, where $\theta=(p,q,d,a_{1\text{---}p},b_{1\text{---}q},c)$ and $S_{x}$ is the stationary output obtained by applying $d^{th}$-order time-recursive differencing on $f(V_x)$. For this estimation, we use:
\begin{eqnarray}\nonumber
 \hat{S}_{F} = a_{1}S_{F-1} + \cdots + a_{p} S_{F-p} + b_{1}E_{F-1}\\ + \cdots +
\left. b_{q}E_{F-q} + E_{F}+ c \right.
\end{eqnarray}
where $E_x$ represents the error terms, a white noise, and obviously the estimated $p,q,d<F$.
\par The initial estimates of the model and hyper- parameters are used to get their final estimates (block-wise) corresponding to each $A_{m,n}$, where the AIC involves the computation of the likelihood $P(s_{m,n-1}/\theta, s_{m,k'}), \forall k'\ \& \ k'\ne n-1$ using the expression in (\ref{eqn}).
\section{Experimental results}
\label{sec:result}
In this section, the effectiveness of our approach is demonstrated using three aerial video datasets, namely Stanford Drone (SD) dataset \cite{robicquet2016learning}, Okutama-action (OA) dataset \cite{barekatain2017okutama} and
 UCF Aerial Action (UCF-AA) dataset \cite{ucf}, and two popular standard anomaly detection datasets, namely the UCSD Pedestrian 2 (Ped2) dataset \cite{ucsd} and the UMN dataset \cite{umn}. The normal behavior in all the three aerial video datasets depicting different scenes is people walking, and therefore, any other activity is considered as anomaly. Considering the standard anomaly detection datasets, the anomaly in the UCSD Ped2 dataset is non-pedestrian in pedestrian path and the anomaly in UMN dataset is sudden running of walking people. We present qualitative and quantitative results of our approach applied to the datasets. Further, our approach is objectively compared with $20$ existing approaches including the state-of-the-art on the standard UCSD Ped2 and UMN datasets.
\subsection{Results and Discussion} 
 We consider $F=10$ and $N=10$ (block size: $10\times 10$) in our approach. As the focus of this paper is on proper prediction for anomaly detection in non-stationary videos, we choose the magnitude of optical flow vector \cite{sun2014quantitative} as the video feature, which is a simple one, to perform anomaly detection and demonstrate the effectiveness of our time-recursive differencing network based prediction driven approach. We consider the value of $\lambda_{f}$ (active blocks will have feature values above this) as the average magnitude of optical flow obtained from the $F$ consecutive non-anomalous frames, to ensure that blocks are not considered active due to background noise. Finally, anomaly is detected by considering the value of $\lambda_{A}$ as $0.01$, which is empirically determined. Further, the aerial videos of SS, OA, UCF-AA are compensated for camera's ego motion using \cite{martius2015vidstab} before anomaly detection.
 \begin{table}[t]
	\begin{center}
	\caption{\label{ucsd-table} Comparisons for UCSD Ped2 Dataset using available quantitative measures}
	\setlength\arraycolsep{3pt} 
	\begin{tabular}{|m{.8cm}|m{.4cm}|m{2.5cm}|m{.8cm}|m{.8cm}|m{.9cm}|}
		\hline
	Model&Date& Method & Frame-level EER $\%$ & Pixel-level EER $\%$& Frame-Level AUC $\%$\\  \hline
		\multirow{2}{*}{AE}&2018&Zhu et al. \cite{zhu2018real} & 7.2 & 14.8 & 97.1	\\
&2015&Xu, Dan, et al. \cite{xu2015learning} & 17 & \textit{n.av.} & 90.8	\\ \hline  
		 \multirow{3}{*}{GAN}  &2017&Ravanbakhsh et al. \cite{ravanbakhsh2017abnormal} & 14 & \textit{n.av.} & 93.5	\\ 
		 &2018&Liu et al. \cite{liu2018future} & \textit{n.av.} & \textit{n.av.} & 95.4	\\ 
 &2020&Ganokratanaa et al. \cite{ganokratanaa2020unsupervised} & 9.4 & 21.8 & 95.5	\\  \hline
\multirow{2}{*}{CNN}&2019&Ionescu et al. \cite{ionescu2019object} & \textit{n.av.} & \textit{n.av.} & 97.8	\\  
&2020&Li et al. \cite{li2020spatial} & 12.7 & \textit{n.av.} & 92.9	\\ \hline
AE+&2018&Singh et al. \cite{singh2018deep} & \textit{n.av.} & \textit{n.av.} & 80.2	\\  
 Conv LSTM &2019&Nawaratne et al. \cite{nawaratne2019spatiotemporal} &8.9 & \textit{n.av.} & 91.1	\\  \hline
	    	     
	  Sparse LSTM& 2019&Zhou et al. \cite{zhou2019anomalynet} & 10.3 & \textit{n.av.} & 94.9	\\ \hline
	   
	SCI-&2019&Xu et al. \cite{xu2019video} & 9.3 &15 &95.3	\\  
	    DNN &2019&Luo et al. \cite{IEEEexample:luo2019video} & \textit{n.av.} &\textit{n.av.} & 92.2	\\  
	    \hline
Gauss. Reg.&2015&Sabokrou et al. \cite{sabokrou2015real} & 19 & 24& \textit{n.av.}\\  \hline
MDT+ &2010&Mahadevan et al. \cite{mahadevan2010anomaly} &25 &\textit{n.av.} & \textit{n.av.}	\\ 
	    	  ARMA  &2013&Li et al. \cite{li2013anomaly} & 18.5 & \textit{n.av.} & \textit{n.av.}	\\ \hline
	    	 AR &2019&Abati et al. \cite{abati2019latent} & \textit{n.av.} & \textit{n.av.} & 95.4	\\ \hline
	    	   Ridge Reg. &2019&Hu et al.  \cite{hu2019two} &14.5 & \textit{n.av.} & 91.9	\\ \hline
	   Ordinal Reg. &2020&Pang et al. \cite{pang2020self} &\textit{n.av.} & \textit{n.av.} & 83.2	\\  \hline
Kernal& 2014&Xu et al. \cite{xu2014video} & 21 & 42& 88.2	\\ 
Learning   &2017&Leyva et al. \cite{leyva2017video} & 19.2 & 36.6& \textit{n.av.}	\\  \hline
	 	   		&&\textbf{Ours} & \textbf{5} & \textbf{11} & \textbf{98.2}	\\  \hline	
	\end{tabular}
	\end{center}
	\hspace{0.1cm} \textit{n.av.} - not available \hspace{0.1cm} Reg. - Regression \hspace{0.1cm} SCI. - Sparse coding inspired
\end{table}
\begin{figure}[t]
	\begin{center}
		\setlength\tabcolsep{1pt}
		\begin{tabular}{ c c}			\includegraphics[width=4.5cm,height=2.5cm]{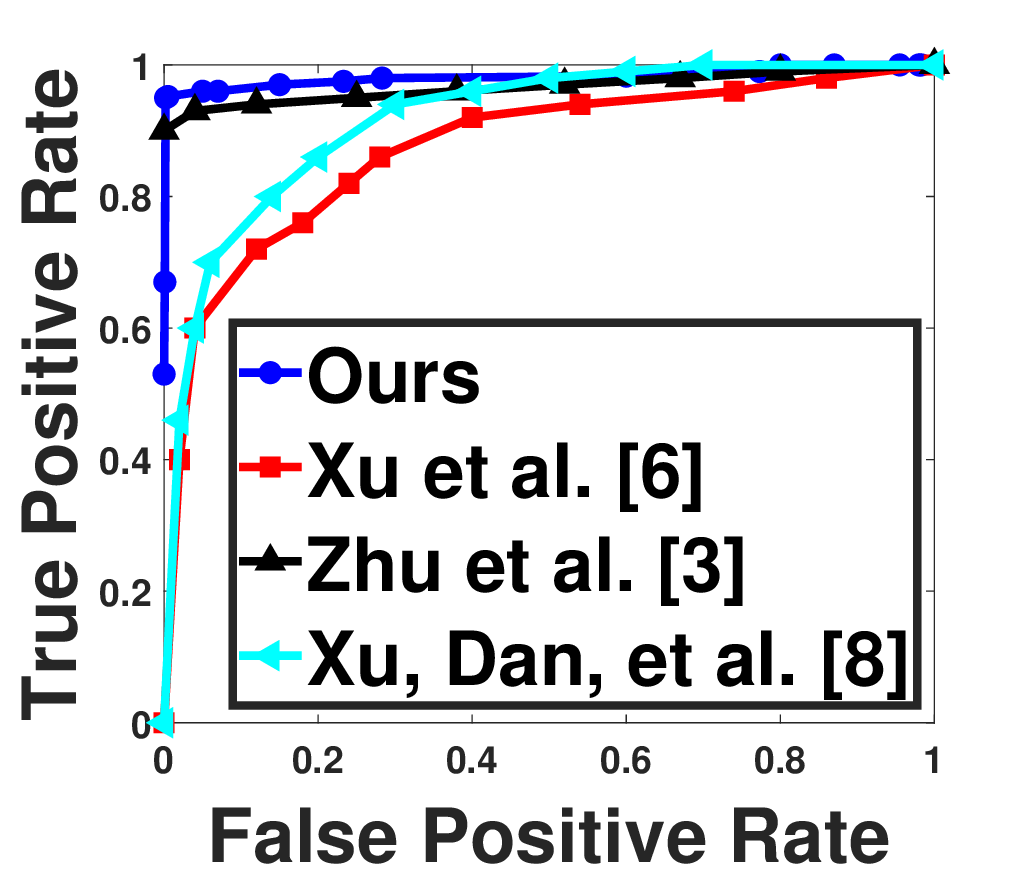} &
			\includegraphics[width=4.5cm,height=2.5cm]{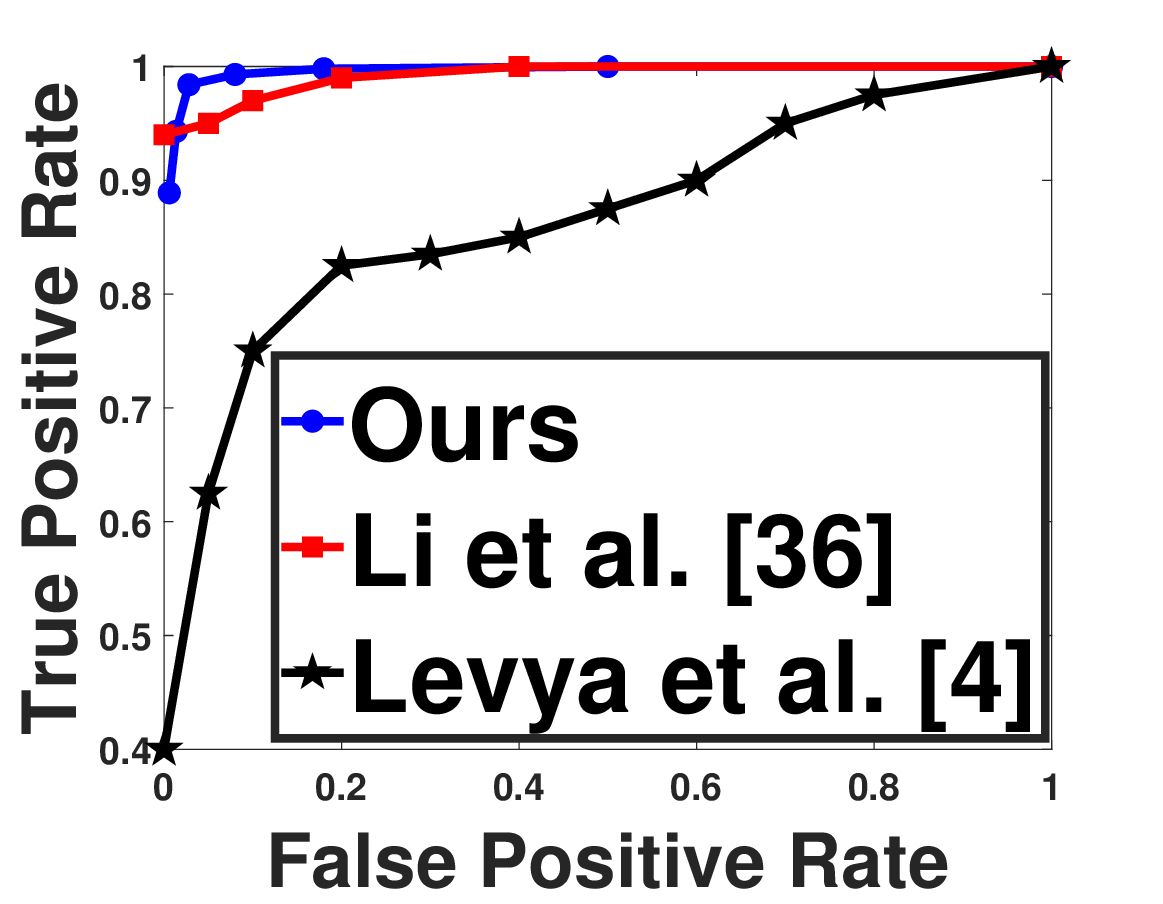}
			\\
		\small{(a)}&\small{(b)}
		\end{tabular}
		\caption{{Frame-level ROC curves for (a) UCSD Ped2, and (b) UMN datasets.}}
		\label{result-roc}
			\end{center}
	\end{figure}
\par Figures~\ref{result_frames_aerial},~\ref{result_frames} and~\ref{consecutive frames} show different qualitative results of our approach on video frames from the 5 datasets. As can be seen, a variety of anomalies in scenes with different backgrounds are detected successfully. Sample results on the SD dataset is given in Figures~\ref{result_frames_aerial}(b) and ~\ref{consecutive frames}(a), where we see that the anomalous activities are successfully detected. Figure~\ref{consecutive frames}(b) shows an example of results on the OA dataset, where we see running people are detected as anomaly in consecutive frames. Figures~\ref{result_frames_aerial}(a),~\ref{consecutive frames}(c) and ~\ref{consecutive frames}(d) show a few results from the the UCF-AA dataset, where people running and vehicle moving are detected as anomalous. Anomaly due to presence of any moving object other than walking people in pedestrian roads are successfully detected in the results shown in Figures~\ref{result_frames} and ~\ref{consecutive frames}(e), which are samples on the UCSD Ped2 dataset. In case of the example results on the UMN dataset given in Figures~\ref{consecutive frames}(f), ~\ref{consecutive frames}(g) and ~\ref{consecutive frames}(h), sudden running of walking people are detected as anomaly.
\begin{table}[t]
	\begin{center}
	\caption{\label{umn-table} Comparisons for UMN Dataset using available quantitative measures}
	\begin{tabular}{|m{2cm}|m{.7cm}|m{3cm}|m{1.1cm}|}
	\hline
Model&Date&Method & AUC $\%$  \\  \hline
	AE&2018&Zhu et al. \cite{zhu2018real} & \textbf{99.7}	\\  \hline
   \multirow{2}{*}{GAN}&2017&Ravanbakhsh et al. \cite{ravanbakhsh2017abnormal} &99	\\ 
     &2020&Ganokratanaa et al. \cite{ganokratanaa2020unsupervised} & 99.6	\\  \hline
	 \multirow{2}{*}{CNN} & 2019&Ionescu et al. \cite{ionescu2019object} & 99.6	\\
	  	&2020&Li et al. \cite{li2020spatial} &98	\\ \hline
	Sparse LSTM	&2019&Zhou et al. \cite{zhou2019anomalynet} & 99.6\\ \hline
	 Gauss. Reg. &2015&Sabokrou et al. \cite{sabokrou2015real} & 99.6	\\ \hline
MDT+ ARMA &2013&Li et al.\cite{li2013anomaly}&99.5 \\   \hline
 Ordinal Reg.	&2020&Pang et al. \cite{pang2020self} & 97.4	\\  \hline
Kernal Learning	&2017&Leyva et al. \cite{leyva2017video} & 88.3	\\  \hline	
	 	   		&&\textbf{Ours} & \textbf{99.7}	\\  \hline
	\end{tabular}
	\end{center}
	\hspace{0.5cm} Reg. - Regression
\end{table}
 \par  To compare our approach quantitatively to the existing methods including the state-of-the-art on UCSD Ped2 dataset, we use equal error rate (EER) both at frame-level, where at least one truly anomalous pixel needs to detected as an anomaly in a frame for the frame to be marked correctly as an anomaly, and at pixel-level, where at least 40 percent of the truly anomalous pixels in a frame needs to be detected as anomalies to mark the frame correctly as an anomaly \cite{sabokrou2015real}. Further, quantitative comparisons on both UCSD Ped2 and UMN datasets are performed using frame-level (a frame anomalous or not) AUC \cite{ionescu2019object}. The quantitative results of our approach and that of the others are shown in Tables~\ref{ucsd-table} and~\ref{umn-table}. As can be seen, our approach performs the best in terms of both EER and AUC when evaluated on the UCSD Ped2 dataset. Our approach also performs as well as the existing when evaluated on the UMN dataset with AUC. The ROC curves shown in Figure.~\ref{result-roc}(a) and Figure.~\ref{result-roc}(b) correspond to some of the AUC results in the two tables, including ours, where our approach is found to perform the best among those shown. We present all the quantitative results of the existing approaches that are available, as reported by the respective papers. In terms of computation speed, our approach works almost at $15$FPS for $240\times 360$ sized video frames on Matlab platform in a 3GHz CPU machine with $16$GB RAM, which is comparable to a few other reported speeds\cite{nawaratne2019spatiotemporal,zhu2018real}. Table~\ref{parameter} shows the variation in performance of our approach with respect to its various parameters, where it is evident that while wide ranges of $F$ and $N$ are suitable, $\lambda_A$ is moderately sensitive.
\par From the above analysis, we find that our approach performs effectively compared to the state-of-the-art even when a simple optical flow based feature is considered. Our approach performs as well as or better than the existing on the 5 varied datasets as shown by both the quantitative and qualitative results. Thus, our time-recursive differencing network, which reduces the effect of non-stationarity of videos (including those captured via aerial remote sensing) during anomaly detection obviously plays a crucial role. The implementation codes of our approach are available at https://github.com/gvp12/video-anomaly-detection.
\vspace{-3mm}
\section{Conclusion}
\label{sec:conclusion}
A video anomaly detection approach has been proposed based on a time-recursive differencing network followed by autoregressive moving average prediction. The anomaly detection has been performed on non-overlapping active blocks of video frames, where spatio-temporal consistency is maintained. Experimental results employing a simple optical flow based feature have demonstrated the effectiveness of the proposed approach on videos, including those captured via aerial remote sensing. The frame- and pixel- level EERs of the approach is 5$\%$ and 11$\%$ on the UCSD Ped2 dataset, respectively, which is lesser than the state-of-the art. AUC values of 98.2$\%$ and 99.7$\%$ have been obtained for the UCSD Ped2 and UMN datasets, respectively, with the former being better than the state-of-the-art and the latter being at par. The qualitative results of our approach on aerial videos from SD, OA, and UCF-AA datasets show that our approach performs successful anomaly detection. In future, use of sophisticated feature learning in conjunction with this new approach will be explored, making it applicable to video data having semantically defined anomalies.
\vspace{-2mm}
\begin{table}[h!]
	\begin{center}
		\caption{\label{parameter} Frame level AUC/EER $\%$ for different values of $\lambda_A$, $F$ and $N$}
		\begin{tabular}{|p{.5cm}|p{1.5cm}|p{.5cm}|p{1.5cm}|p{.5cm}p{1.5cm}}
			\hline
			$\lambda_A^{*}$ & AUC/EER $\%$ & $F^{\ddagger}$ &  AUC/EER $\%$& \multicolumn{1}{c|}{$N^{\dagger}$} & \multicolumn{1}{c|}{AUC/EER $\%$} \\  \hline
			.001 & 74.0/33 & 3 & 97.9/8& \multicolumn{1}{c|}{5} & \multicolumn{1}{c|}{96.8/7}\\\hline
			.005 & 78.2/30 &5 & 98.0/6 & \multicolumn{1}{c|}{\textbf{10}} & \multicolumn{1}{c|}{\textbf{98.2/5}}\\ \hline
			\textbf{.01} & \textbf{98.2/5} & \textbf{10} & \textbf{98.2/5}& \multicolumn{1}{c|}{20} & \multicolumn{1}{c|}{98.2/5} \\ \hline
			.1 & 94.1/7 & 15 & 98.2/5&   & \\ 
			\cline{1-4}
			1 & 93.0/8 & 20 & 98.2/5&& \\ 
			\cline{1-4}
		\end{tabular}
	\vspace{-1mm}
	\end{center}
	\hspace{0.5cm} $^*$ $F$=10, $N$=10 \hspace{0.5cm} $^\ddagger$ $\lambda_A$=.01, $N$=10 \hspace{0.5cm} $^\dagger$  $\lambda_A$=.01, $F$=10
\end{table}
\vspace{-5mm}
\section{Aknowledgement}
Debashis Sen acknowledges the Science and Engineering Research Board (SERB), India for its assistance.
\vspace{-3mm}

\end{document}